\documentclass[doc]{apa6}
\usepackage[utf8]{inputenc}

\usepackage[round]{natbib}

\usepackage{multirow}
\usepackage{booktabs}
\usepackage{graphicx}
\usepackage{textcomp}
\usepackage{xcolor}
\usepackage{tikz}
\usepackage{subcaption}

\title{Generalisation in Neural Networks Does not Require Feature Overlap}
\shorttitle{Generalisation in Neural Networks}

\twoauthors{Jeff Mitchell (Corresponding Author)}{Jeffrey S. Bowers} 
\leftheader{Mitchell \& Bowers}
\twoaffiliations{School of Computing, Edinburgh Napier University, 10 Colinton Road, Edinburgh, UK. }{School of Psychological Science, University of, Bristol,  12 Priory Road, Bristol, UK.  }
\authornote{{\tt j.mitchell@napier.ac.uk} , {\tt j.bowers@bristol.ac.uk} \paragraph{Declarations of interest:} none}

\abstract{
    That shared features between train and test data are required 
    for generalisation in artificial neural networks has been 
    a common assumption of both proponents and critics of 
    these models. 
    Here, we show that convolutional architectures avoid this 
    limitation by applying them to two well known challenges, 
    based on learning the identity function and learning rules 
    governing sequences of words. 
    In each case, successful performance on the test set 
    requires generalising to features that were not present 
    in the training data, which is typically not feasible 
    for standard connectionist models.
    However, our experiments demonstrate that neural 
    networks can succeed on such problems when they 
    incorporate the weight sharing employed by 
    convolutional architectures. 
    In the image processing domain, such architectures 
    are intended to reflect the symmetry under spatial 
    translations of the natural world that such images depict. 
    We discuss the role of symmetry in the two tasks 
    and its connection to generalisation.}

\date{}

\begin{document}

\maketitle

\section{Introduction}

In this article, we examine two problems 
\citep{marcus1998,marcusetal1999}
designed to highlight the shortcomings of 
connectionist models that rely on 
generalisation based on featural similarity or overlap 
and describe solutions to these long-standing\footnote{In the eyes of its creator, the identity learning task has remained an unresolved problem for neural network models for over two decades \citep{marcus1998,marcus2020}.} challenges. 
The identity and rule learning tasks themselves are described 
in more detail in Sections \ref{id_sec} and \ref{rl_sec}, 
but the basic idea in each case is fairly simple. 
Both tasks require generalising to test data that contains 
elements - digits or words - that were not seen during training.

Prior research on such tasks has often relied on 
modifications to the representations of these elements, 
via careful input coding or appropriate pre-training,
that introduce shared features as a means 
to achieve generalisation to the task test sets 
\citep[e.g.,][]{negishi,seidenberg,sirois}. 
In fact, it has commonly been asserted that 
generalisation in neural networks requires such featural overlap 
\citep{alhamazuidema2019,marcus2001,mcclellandplaut1999} 
and this characteristic is frequently invoked as a benefit of 
distributed representations \citep{hinton86,farrelllewandowsky2000}.
Here, we show that these models are capable of 
generalising to test data that lacks any features in common 
with their training data by exploiting symmetries 
relevant to the task. 
Architecturally, this is achieved using convolutional 
\citep{neocognitron,Lecun98gradient-basedlearning}
layers, a common component in deep learning approaches 
to image categorisation, 
and our solutions repurpose their spatial symmetries 
to serve the needs of each problem. 

While the computational processes involved in 
convolution are probably not biologically or cognitively 
plausible in their particular detail, 
we nonetheless wish to argue that symmetry, 
considered more abstractly, is 
a useful concept in thinking about how generalisation 
beyond similarity is possible. 
By symmetry, we mean a transformation, 
or set of transformations, that preserve some relevant 
characteristics of a task. 
For example, in a task such as object recognition, 
a particular dog or face is equivalent wherever it occurs 
within an image, and this can be thought of as 
a symmetry under spatial translations.
Thus, ideally, a neural network model of object recognition 
should be able to generalise from a dog seen during training 
in the bottom right of an image to the same dog at test time 
in the top left, even though there is no overlap in the 
set of pixels involved. 
This implies that spatial translation, 
e.g. from bottom right to top left, should be a symmetry of the 
function computed by the network, 
and convolutional networks implement this by 
learning feature weights that are shared across the image 
and then pooling these shared features into 
larger and larger receptive fields 
(described in more detail in Appendix \ref{app_cnn}). 

Our experiments show that symmetry has a wider application, 
beyond just spatial translation within images. 
In particular, we apply this approach to 
learning the identity function on binary digits 
\citep{marcus1998} and to 
rule learning on sequences of words
\citep{marcusetal1999}. 
Successful performance on these tasks
requires generalisation to unseen features, 
i.e. novel digits or words. 
In each case, we analyse the structure of the domain 
to identify a symmetry that is relevant to the task 
and impose this on the network in terms of 
a convolutional architecture. 
Concretely, this results in weight sharing between the 
seen and unseen features, 
i.e. across digit positions or syllables, 
allowing effective generalisation to the test instances. 

\cite{marcus2001} argues that such generalisation 
to novel features is not possible for connectionist 
models due to what he calls \emph{training independence}. 
Roughly, he assumes that the weights within these 
networks are learned independently, with the consequence 
that no updates are made to the connections 
attached to any unit which is inactive during training. 
Although they are critical of the 
conclusions drawn by \cite{marcusetal1999} 
from the rule learning experiments, 
\cite{mcclellandplaut1999} nonetheless affirm 
that generalisation in neural networks 
depends on overlapping patterns of activity 
somewhere in the network. 
Furthermore, in a review of the literature 
surrounding those experiments, \cite{alhamazuidema2019} 
also assume that generalisation will 
depend on the overlap between vectors 
in training and test. 

Thus, the requirement of featural overlap 
to achieve generalisation has been a basic assumption 
made by both proponents and critics of connectionist 
models of cognition. 
However, convolutional neural nets are designed precisely 
to overcome this limitation in relation to generalisation 
to unseen positions, and this ability  has been a key factor 
in their widespread adoption in deep image recognition 
models. 
The mechanism by which they achieve this is simply the sharing 
of the same feature detectors across all positions in the 
image, which avoids the independence of weights 
criticised by \cite{marcus2001}. 
More abstractly, this weight sharing can be understood as
an implementation of the requirement for the network's 
behaviour to be symmetric under spatial translations. 
Thus, in addition to generalisation being possible between 
inputs related by featural similarity, we can also obtain 
generalisation between inputs related by a symmetry. 

In the following sections, 
we explore how this idea can be applied 
to the identity and rule learning challenges, 
demonstrating convolutional architectures that achieve 
effective generalisation to the unseen digits and syllables. 
In each experiment, we compare a network with an appropriate 
symmetry constraint to the same network without the constraint.

\section{Identity Learning}\label{id_sec}

\begin{table}[t]
\begin{center}
    \begin{tabular}{l c c c c}
    \toprule
    & ~ & Input & ~ & Output \\
    \midrule 
    \multirow{6}{*}{Training Set} & & 00010 & & 00010 \\
    & & 00100 & & 00100 \\
    & & 00110 & & 00110 \\
    & & 01000 & & 01000 \\
    & & 01010 & & 01010 \\
    & & \ldots & & \ldots \\
    \midrule 
    \multirow{6}{*}{Test Set} & & 00011 & & 00011 \\
    & & 00101 & & 00101 \\
    & & 00111 & & 00111 \\
    & & 01001 & & 01001 \\
    & & 01011 & & 01011 \\
    & & \ldots & & \ldots \\
    \bottomrule
    \end{tabular}
\end{center}
\caption{Examples from the training and test sets used in the 
identity learning task. Each instance consists of the same 
5-digit binary numeral in both input and output, 
representing the identity function. 
However, the training set consists of only the even numbers, 
and the test set consists of the odd numbers. In other words, 
successful generalisation requires handling 
the digit \textbf{1} in a an unseen position.}\label{id_data}
\end{table}

\cite{marcus1998} introduced his identity learning task as 
an example of a situation in which humans are able to 
generalise outside their training space. 
The task simply requires learning the identity function, 
from inputs to outputs, for binary digits. 
In other words, the desired outcome is for the learner 
to copy whatever arrives in the input to the output. 

However, the training data only contains even numbers, 
in which the final digit is always zero, but generalising 
effectively to the test set requires handling odd numbers, 
in which the final digit is one. 
Examples of the training and test data 
are shown in Table \ref{id_data}.
\cite{marcus1998} reports that humans typically 
learn the identity function successfully, 
copying the odd numbers without problem, 
even though they contain a feature that was not 
present during training. 

In contrast, feedforward networks ordinarily fail to generalise 
in this way, precisely because the odd numbers in the 
test set contain an unseen feature. 
This points to an important difference 
in the biases that guide learning in 
people and artificial neural networks. 
Whereas humans find it natural to apply 
the same copying operation to all digit positions, 
standard feedforward networks have no such preference. 

The lack of this bias in these models is not surprising, 
as they are designed to learn 
general input-output associations, and in many 
of the tasks they are applied to there is no identity function. 
For example, the inputs in the MNIST dataset \citep{mnist}
are $28 \times 28$ pixel arrays and the outputs are 
ten digit categories. 
In this case, it is not possible for the input and output 
units to carry identical patterns of activity as 
the numbers of units in each is not even the same. 
There is therefore no unique, well-defined identity function 
between the pixel space and the digit category space.
Even in the case where input and output 
have the same dimensionality, they remain distinct spaces 
and there need not be a single, unambiguous identity function. 
For example, if our task is translation and the inputs are 
constructed from a vocabulary of 100k French words, 
while the outputs use a vocabulary of 100k English words, 
then there is  again no well defined identity function 
between these spaces, despite their being the same size. 

For people presented with the challenge 
described by \cite{marcusetal1999}, 
a relation of identity between inputs and outputs 
will be immediately apparent. 
Moreover, they will also immediately grasp that 
both inputs and outputs are made up of strings of digits 
and that there is a correspondence between digit positions 
in the input and digit positions in the output. 
It is this intrinsic structure in the task 
that makes it tractable for humans. 
In contrast, the standard fully connected architecture 
is indifferent to this structure, 
and for such a network the task consists of identifying 
one unexceptional one-to-one mapping from among many.

While humans find Marcus's task easy because 
the identity function is obvious to them 
and stands out in some way, 
we can also imagine other tasks which lack 
the relevant correspondences between symbols 
and positions in input and output.
In particular, we could efface this structure by replacing 
the digits \textbf{0} and \textbf{1} 
with arbitrary symbols at each position 
and permuting the order of those symbols in the output. 
So, for example, an input of \textbf{!h*7@} 
might map onto the output \textbf{8j?:y}. 
In this case, learning such an arbitrary connection between 
input and output and then generalising 
to an unseen symbol is going to be much more difficult, 
and more similar to the problem that a standard connectionist 
network is trying to solve.

Just as we can make the inherent structure in the problem 
more obscure for the biological neural systems of human beings, 
we can also try to modify the architecture 
of an artificial neural network to exploit that structure 
more effectively. 
In particular, if we want the network to be able to 
apply the same operation to the final digit that it applied 
to all the others, then its design ought to reflect the fact 
that the same digit can occur in multiple positions 
and that there is a correspondence between 
positions in the input and positions in the output. 

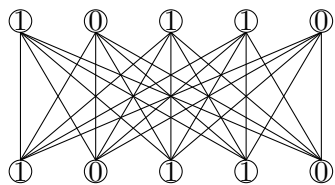
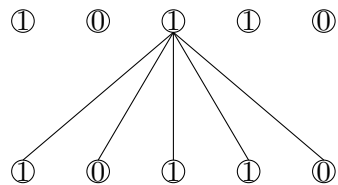
\begin{figure*}[t]
  \centering
  \begin{subfigure}[t]{0.45\textwidth}
  \centering
\begin{tikzpicture}[draw=black!50, node distance=\layersep, scale=0.1]

\draw [black] (0,0) circle (1.5cm) node {1};
\draw [black] (10,0) circle (1.5cm) node {0};
\draw [black] (20,0) circle (1.5cm) node {1};
\draw [black] (30,0) circle (1.5cm) node {1};
\draw [black] (40,0) circle (1.5cm) node {0};

\draw [black] (0,20) circle (1.5cm) node {1};
\draw [black] (10,20) circle (1.5cm) node {0};
\draw [black] (20,20) circle (1.5cm) node {1};
\draw [black] (30,20) circle (1.5cm) node {1};
\draw [black] (40,20) circle (1.5cm) node {0};

\draw [black] plot coordinates {(0,1.5) (0,18.5)};
\draw [black] plot coordinates {(10,1.5) (0,18.5)};
\draw [black] plot coordinates {(20,1.5) (0,18.5)};
\draw [black] plot coordinates {(30,1.5) (0,18.5)};
\draw [black] plot coordinates {(40,1.5) (0,18.5)};

\draw [black] plot coordinates {(0,1.5) (10,18.5)};
\draw [black] plot coordinates {(10,1.5) (10,18.5)};
\draw [black] plot coordinates {(20,1.5) (10,18.5)};
\draw [black] plot coordinates {(30,1.5) (10,18.5)};
\draw [black] plot coordinates {(40,1.5) (10,18.5)};

\draw [black] plot coordinates {(0,1.5) (20,18.5)};
\draw [black] plot coordinates {(10,1.5) (20,18.5)};
\draw [black] plot coordinates {(20,1.5) (20,18.5)};
\draw [black] plot coordinates {(30,1.5) (20,18.5)};
\draw [black] plot coordinates {(40,1.5) (20,18.5)};

\draw [black] plot coordinates {(0,1.5) (30,18.5)};
\draw [black] plot coordinates {(10,1.5) (30,18.5)};
\draw [black] plot coordinates {(20,1.5) (30,18.5)};
\draw [black] plot coordinates {(30,1.5) (30,18.5)};
\draw [black] plot coordinates {(40,1.5) (30,18.5)};

\draw [black] plot coordinates {(0,1.5) (40,18.5)};
\draw [black] plot coordinates {(10,1.5) (40,18.5)};
\draw [black] plot coordinates {(20,1.5) (40,18.5)};
\draw [black] plot coordinates {(30,1.5) (40,18.5)};
\draw [black] plot coordinates {(40,1.5) (40,18.5)};

\end{tikzpicture}
  \caption{The unconstrained network has an individual weight for each pair of input and output nodes, which must be learned separately.}\label{id_full_fig}
  \end{subfigure}  
  \hfill
  \begin{subfigure}[t]{0.45\textwidth}
  \centering
\begin{tikzpicture}[draw=black!50, node distance=\layersep, scale=0.1]

\draw [black] (0,0) circle (1.5cm) node {1};
\draw [black] (10,0) circle (1.5cm) node {0};
\draw [black] (20,0) circle (1.5cm) node {1};
\draw [black] (30,0) circle (1.5cm) node {1};
\draw [black] (40,0) circle (1.5cm) node {0};

\draw [black] (0,20) circle (1.5cm) node {1};
\draw [black] (10,20) circle (1.5cm) node {0};
\draw [black] (20,20) circle (1.5cm) node {1};
\draw [black] (30,20) circle (1.5cm) node {1};
\draw [black] (40,20) circle (1.5cm) node {0};

\draw [black] plot coordinates {(0,1.5) (20,18.5)};
\draw [black] plot coordinates {(10,1.5) (20,18.5)};
\draw [black] plot coordinates {(20,1.5) (20,18.5)};
\draw [black] plot coordinates {(30,1.5) (20,18.5)};
\draw [black] plot coordinates {(40,1.5) (20,18.5)};

\end{tikzpicture}
  \caption{A single filter in the convolutional network. The same weights are replicated across the other positions in the network.}\label{id_conv_fig}
  \end{subfigure} 
  \caption{The connection weights of the two architectures applied to the identity learning task.}
\end{figure*}

In fact, this is exactly the problem 
that convolutional layers are designed to solve. 
Effective image processing requires that the same objects 
or features can be recognised wherever they occur in an image 
without losing the knowledge of where they occur. 
A convolutional layer addresses this by applying the same set of filters across the whole of an image, producing 
an equivalent bank of channels at each position. 
This introduces both the notion of the same structure being 
able to occur in multiple positions, and also a correspondence 
between input positions and output positions. 

In mechanistic terms, this is achieved by sharing the 
same connection weights from inputs to outputs 
at each position within the network. 
However, at a more abstract level we can see this as 
an implementation of translational symmetry 
in the function computed by the network. 
In particular, if that function, $y=f(x)$,
maps an input image, $x$, to a feature map, $y$, 
then we can consider the result when the function is 
applied to a translation, $Tx$, of the original image. 
Convolutional layers are \emph{equivariant} under these 
translations: $f(Tx)=Ty$. 
That is, the output from a translated image is itself 
a translated version of the feature map for the original image. 
Or, more concretely, if a set of feature detectors 
are active in response to a structure in the top right 
of an image, then translating that structure diagonally across 
the image into the opposite corner will result in an 
equivalent set of feature detectors firing in the bottom left. 

This translational symmetry enables the convolutional layer 
to automatically generalise from features seen during training 
in one position to novel, unseen positions at test time. 
Moreover, this happens without any featural overlap between 
the training and test inputs, as the features are present 
at different positions. 
What connects the seen and unseen inputs is not their 
pixel similarity but instead the fact that they can be mapped 
onto each other by a symmetry transformation. 

In terms of the problem proposed by \cite{marcus1998}, 
the inputs and outputs consist of sequences of digits, 
in which the same digit, e.g. \textbf{0} or \textbf{1},
can occur in multiple positions, 
and those positions have an intrinsic order in both 
the inputs and outputs. 
This implies that we can meaningfully talk about moving a 
digit from one position to another, 
and that there is a correspondence between such motions 
in the input and the output. 
Convolution imposes this structure on the architecture 
and makes the behaviour of the net symmetric 
under these translations. 

\subsection{Experiment 1}

We consider two architectures: a standard, fully connected 
network and a convolutional network, 
both with 5 inputs and outputs 
and both consisting of a single layer without hidden units. 
Thus, the only difference between these models 
is the constraint of translational symmetry 
imposed on the latter. 
The set of functions computable by the convolutional network 
is therefore a strict subset of those computable 
by the unconstrained network. 
This experiment investigates how this constraint 
affects learning and generalisation 
in the identity learning task. 

The data consist of all five-digit binary numbers, 
ranging from zero to thirty one, 
and all inputs and outputs are equal. 
The training data consists exclusively of even numbers, 
in which the least significant digit is zero, 
and the test data exclusively of odd numbers, 
in which the least significant digit is one. 

The experiments are implemented in Tensorflow \citep{abadi}, 
using batch gradient descent  
over the full training set of 16 instances. 
The squared error loss function is minimised 
for 1,000 epochs, and then performance is evaluated 
over the entire training and test sets. 
An output is treated as correct if all its digits 
are predicted correctly, based on a cutoff of 0.5 
to discretise units activities into binary values. 
We report average accuracy over 100 training runs 
with random initialisations.

\begin{table}[t]
  \begin{center}
  \begin{tabular}{lrr}
    \toprule
    
    Architecture & Training Accuracy &  Test Accuracy \\
    \midrule 
    Unconstrained & 100\% &  12\% \\
    Convolutional & 100\% & 100\% \\
    \bottomrule
  \end{tabular}
  \caption{Accuracies on the training and test sets of the identity learning task, with and without the constraint of translational symmetry.}\label{id_res}
  \end{center}
\end{table}

The results are reported in Table \ref{id_res}. 
The pattern of performance seen 
for the standard, unconstrained network 
in the first row indicate that, 
while learning succeeded  
over the training data, generalisation to the test 
set was not as successful. 
This was precisely because no updates were made to the 
weights attached to the inactive digit. 
In contrast, the accuracies in the second row of the table 
indicate perfect generalisation
for the convolutional architecture. 
Here, weight sharing allowed learning to occur 
across all positions, even the inactive one.

\subsection{Discussion}

Our experiments showed that a convolutional network 
was able to  
generalise successfully to unseen digits 
without featural overlap between training and test instances. 
Thus, one potential solution to the identity learning challenge 
\citep{marcus1998} turns out to be a standard architectural 
design that predates it by almost two decades 
\citep{neocognitron}.
In concrete terms, weight sharing across 
digit positions avoids the problem of \emph{training 
independence} identified by \cite{marcus1998}, 
allowing what is learned at seen positions 
to be transferred to unseen positions. 
More abstractly, symmetry provides a notion of \emph{sameness} 
that goes beyond featural overlap, 
allowing us to apply the \emph{same} operation to 
features that were unseen during training. 
Here, we identified translational symmetry as reflecting 
the intrinsic structure of the problem, in terms of 
inputs and outputs being sequences of binary digits. 

In the task, this symmetry relates to the fact that 
the same digit can occur in multiple positions, 
and the equivariance of the convolutional layer 
provides the required correspondence between 
input and output spaces. 
More generally, a symmetry under movement of elements 
between positions may be relevant whenever we are dealing with 
structures in which the same components can be rearranged 
in multiple configuration, such as list or stack 
memory structures. 

However, unless these ideas can be shown to be 
applicable in a broader range of tasks, then it might 
be reasonable to treat the approach described here 
as merely a convenient trick or hack. 
In particular, it would be desirable to show that 
considerations of symmetry are relevant beyond 
simple spatial translations and without the 
requirement for the input and output spaces 
to share a common structure.

\section{Rule Learning}\label{rl_sec}

\begin{table}[t]
\begin{center}
    \begin{tabular}{l c c c c}
    \toprule
    & ~ & Input & ~ & Output \\
    \midrule 
    \multirow{5}{*}{Training Set} & & ga ti ga & & ABA \\
    & & li gi li & & ABA \\
    & & ga ti ti & & ABB \\
    & & li na na & & ABB \\
    & & \ldots & & \ldots \\
    \midrule 
    \multirow{4}{*}{Test Set} & & wo fe wo & & ABA \\ 
    & & de ko de & & ABA \\
    & & wo fe fe & & ABB \\ 
    & & de ko ko & & ABB \\
    \bottomrule
    \end{tabular}
\end{center}
\caption{Examples from the training set and the full test set 
used in the rule learning task. Each input is a sequence 
of three words which are categorised at the output into 
the structures ABA and ABB. The words used to construct 
the sequences in the test set are entirely disjoint from 
those in the training set. In other words, 
successful generalisation requires handling the unseen 
words appropriately.}\label{rl_data}
\end{table}

In the previous task, effective generalisation required 
applying the same rule to a known digit when it occurred 
in a new position. 
The connection to image processing and the need to 
recognise known features in new positions was reasonably 
straightforward.
In this section we will consider a task based on 
word sequences rather than digits, and which requires 
generalisation to new words rather than new positions. 
While translation between positions is no longer 
the relevant symmetry, we will nonetheless show that 
convolution can also be used to solve this problem.

This rule learning task involves learning to differentiate 
between structures of the form ABA and ABB, 
and is based on the the experiments with infants described by 
\cite{marcusetal1999}. 
Examples of the task data can be seen in Table \ref{rl_data}.
Each input to the network is a sequence of three words, 
such as \emph{ga ti ga} or \emph{ga ti ti},
and the outputs are binary labels  
indicating the ABA and ABB categories. 
Crucially, however, the sequences in the test data are 
constructed from words that were not seen during training, 
e.g. \emph{wo fe wo} or \emph{wo fe fe}.
In other words, the network must learn 
in a sufficiently abstract manner to be able to 
apply the same rule to both seen and unseen words. 

In their experiments, \cite{marcusetal1999} familiarised 
7-month-old infants with sequences from one of the categories 
and then subsequently observed their responses to 
sequences containing novel words from both categories. 
Overall, the infants showed more interest in stimuli 
from the unfamiliar category, indicating they were able 
to generalise their experience to words they did not hear 
during the familiarisation phase. 
\cite{marcusetal1999} were unable to obtain this form of 
generalisation from a neural network trained on the same data, 
because the backpropagation learning rule did not 
update the weights associated with unseen words during training. 
Numerous studies have investigated a variety of aspects of 
these experiments, 
including how phonetic forms are represented \citep{negishi}, 
the nature of the familiarisation process \citep{sirois} 
and the knowledge already acquired by the infants before 
the experiments \citep{seidenberg}.

Setting aside other psychological aspects, 
we focus here on the core computational question of 
whether a neural network can be endowed with the ability 
to learn about structure in a way that allows it to 
generalise to instantiations of that structure constructed 
from novel elements. 
To this end, we frame the problem as 
a supervised learning task in which each of the three 
words in the input sequence are represented 
by a one-hot vector, thus ensuring there is 
no featural overlap between train and test, 
and the output consists of a pair of units 
representing the categories ABA and ABB. 
The data used in our experiments are taken from 
the \cite{marcusetal1999} paper, and we evaluate the models 
in terms of their ability to generalise to test inputs 
that are constructed from words not seen during training. 
As a result of the localist coding of words we have employed 
the training and test inputs share no active units in common, 
and so successful generalisation requires that the network 
is able to generalise outside its training space.

However, the architecture used in the previous experiment 
is not relevant to the current task. 
In that task, inputs and outputs lay in the same space, 
which permitted us to impose a constraint based on 
symmetry under translations acting on both spaces. 
Here, the output space is entirely distinct 
from the input space and, in particular, there is 
no meaningful identity function between them. 
Instead, the pair of output units are intended to categorise 
the structure of the input word sequence. 
Specifically, we want to learn to categorise  
these sequences based purely on their structure, 
independently of the particular words 
instantiating that structure. 

In other words, the outputs of the network should be invariant 
to word substitutions that leave the structure of 
the sequence unchanged. 
For example, replacing \emph{ti} with \emph{li} 
in \emph{ga ti ti} to produce \emph{ga li li} 
does not change the structure of the sequence 
and so both sequences should be categorised in the same way. 
Thus, such substitutions should be symmetries of the network 
in terms of having no impact on the outputs. 
In this way, we can ensure that the categorisation of 
sequences is driven by their structure and not by 
the particular words seen during training.

This is comparable to the invariance under spatial translation 
that visual object recognition systems seek to achieve. 
In that case, categorisation should be driven purely 
by the shape of an object and not by the position 
it occurred in during training. 
Convolutional architectures address this by combining 
convolutional layers, which are \emph{equivariant} under translations, 
with pooling layers, which are \emph{invariant} under translations.\footnote{In fact, as described in Appendix \ref{app_cnn}, the architectures applied in practice to image recognition tasks often fall short of full translation in variance \citep{blything,azulayweiss2019,biscionebowers2020}.}
Specifically, the output of a pooling function, 
$y=p(x)$ remains unchanged when the input, $x$, 
is translated within its receptive field: 
$p(x)=p(Tx)$. 
Commonly, a fixed pooling function, 
such as maximum or average, is used to combine the signals 
across a range of positions, creating larger receptive fields, 
which then feed into the next convolutional layer, 
until a final global pooling layer is applied 
across the whole image.
That is, in simple terms, a convolutional layer 
produces a feature for every position in the input 
and a global pooling layer aggregates these features 
to give a single position independent output. 

Whereas translations are the relevant symmetry for 
object recognition, here we wish to make our network 
invariant to the substitution of one word with another. 
We will, nonetheless, be able to employ the same 
convolution and pooling layers in constructing a solution. 
In particular, by treating word types as if they were positions 
in an image, word substitutions become analogous to 
movement between these positions. 
A convolutional layer can then extract the same structural 
features for each word type, and a pooling layer can 
reduce this to a single pair of output categories 
which are invariant to the word types involved. 

\subsection{Experiment 2}

\begin{figure*}[t]
\begin{center}
\begin{tikzpicture}[scale=0.05]

\draw [white] (5,30) rectangle ++(15,10) node[pos=.5,black] {wo};
\draw [white] (5,40) rectangle ++(15,10) node[pos=.5,black] {fe};
\draw [white] (5,50) rectangle ++(15,10) node[pos=.5,black] {wo};

\draw [black] (5,30) rectangle ++(15,30);
\draw [black,->] plot coordinates {(20,45) (30,45)};

\draw [white] (30,20) rectangle ++(10,10) node[pos=.5,black] {ga};
\draw [white] (40,20) rectangle ++(10,10) node[pos=.5,black] {ti};
\draw [white] (50,20) rectangle ++(10,10) node[pos=.5,black] {wo};
\draw [white] (60,20) rectangle ++(10,10) node[pos=.5,black] {na};
\draw [white] (70,20) rectangle ++(10,10) node[pos=.5,black] {gi};
\draw [white] (80,20) rectangle ++(10,10) node[pos=.5,black] {la};
\draw [white] (90,20) rectangle ++(10,10) node[pos=.5,black] {li};
\draw [white] (100,20) rectangle ++(10,10) node[pos=.5,black] {fe};
\draw [white] (110,20) rectangle ++(10,10) node[pos=.5,black] {ko};
\draw [white] (120,20) rectangle ++(10,10) node[pos=.5,black] {ni};
\draw [white] (130,20) rectangle ++(10,10) node[pos=.5,black] {ta};
\draw [white] (140,20) rectangle ++(10,10) node[pos=.5,black] {de};
\draw [black] (30,30) rectangle ++(10,10);
\draw [black] (40,30) rectangle ++(10,10);
\draw [black,fill=black] (50,30) rectangle ++(10,10);
\draw [black] (60,30) rectangle ++(10,10);
\draw [black] (70,30) rectangle ++(10,10);
\draw [black] (80,30) rectangle ++(10,10);
\draw [black] (90,30) rectangle ++(10,10);
\draw [black] (100,30) rectangle ++(10,10);
\draw [black] (110,30) rectangle ++(10,10);
\draw [black] (120,30) rectangle ++(10,10);
\draw [black] (130,30) rectangle ++(10,10);
\draw [black] (140,30) rectangle ++(10,10);
\draw [black] (30,40) rectangle ++(10,10);
\draw [black] (40,40) rectangle ++(10,10);
\draw [black] (50,40) rectangle ++(10,10);
\draw [black] (60,40) rectangle ++(10,10);
\draw [black] (70,40) rectangle ++(10,10);
\draw [black] (80,40) rectangle ++(10,10);
\draw [black] (90,40) rectangle ++(10,10);
\draw [black,fill=black] (100,40) rectangle ++(10,10);
\draw [black] (110,40) rectangle ++(10,10);
\draw [black] (120,40) rectangle ++(10,10);
\draw [black] (130,40) rectangle ++(10,10);
\draw [black] (140,40) rectangle ++(10,10);
\draw [black] (30,50) rectangle ++(10,10);
\draw [black] (40,50) rectangle ++(10,10);
\draw [black,fill=black] (50,50) rectangle ++(10,10);
\draw [black] (60,50) rectangle ++(10,10);
\draw [black] (70,50) rectangle ++(10,10);
\draw [black] (80,50) rectangle ++(10,10);
\draw [black] (90,50) rectangle ++(10,10);
\draw [black] (100,50) rectangle ++(10,10);
\draw [black] (110,50) rectangle ++(10,10);
\draw [black] (120,50) rectangle ++(10,10);
\draw [black] (130,50) rectangle ++(10,10);
\draw [black] (140,50) rectangle ++(10,10);

\draw [black,->] plot coordinates {(90,62.5) (90,77.5)};

\draw [black,fill=white] (90,70) circle (60pt);
\draw (90,70) node {$*$};

\draw [black,fill=black] (165,62.5) rectangle ++(5,5);
\draw [black] (165,67.5) rectangle ++(5,5);
\draw [black,fill=black] (165,72.5) rectangle ++(5,5);

\draw [black,fill=black] (172.5,62.5) rectangle ++(5,5);
\draw [black,fill=black] (172.5,67.5) rectangle ++(5,5);
\draw [black] (172.5,72.5) rectangle ++(5,5);

\draw [white] (165,50) rectangle ++(12.5,10) node[pos=.5,black] {Filters};

\draw [black,->] plot coordinates {(162.5,70) (92,70)};

\draw [black] (30,80) rectangle ++(10,10);
\draw [black] (40,80) rectangle ++(10,10);
\draw [black,fill=black] (50,80) rectangle ++(10,10);
\draw [black] (60,80) rectangle ++(10,10);
\draw [black] (70,80) rectangle ++(10,10);
\draw [black] (80,80) rectangle ++(10,10);
\draw [black] (90,80) rectangle ++(10,10);
\draw [black] (100,80) rectangle ++(10,10);
\draw [black] (110,80) rectangle ++(10,10);
\draw [black] (120,80) rectangle ++(10,10);
\draw [black] (130,80) rectangle ++(10,10);
\draw [black] (140,80) rectangle ++(10,10);

\draw [black] (30,90) rectangle ++(10,10);
\draw [black] (40,90) rectangle ++(10,10);
\draw [black] (50,90) rectangle ++(10,10);
\draw [black] (60,90) rectangle ++(10,10);
\draw [black] (70,90) rectangle ++(10,10);
\draw [black] (80,90) rectangle ++(10,10);
\draw [black] (90,90) rectangle ++(10,10);
\draw [black] (100,90) rectangle ++(10,10);
\draw [black] (110,90) rectangle ++(10,10);
\draw [black] (120,90) rectangle ++(10,10);
\draw [black] (130,90) rectangle ++(10,10);
\draw [black] (140,90) rectangle ++(10,10);

\draw [black,->] plot coordinates {(60,105) (85,115)};
\draw [black,->] plot coordinates {(90,105) (90,115)};
\draw [black,->] plot coordinates {(120,105) (95,115)};

\draw [black,fill=black] (85,120) rectangle ++(10,10);
\draw [black] (85,130) rectangle ++(10,10);

\draw [white] (100,120) rectangle ++(10,10) node[pos=.5,black] {ABA};
\draw [white] (100,130) rectangle ++(10,10) node[pos=.5,black] {ABB};

\draw [white] (200,125) rectangle ++(50,10) node[pos=.5,black] {Softmax};
\draw [white] (200,110) rectangle ++(50,10) node[pos=.5,black] {Max-Pooling};
\draw [white] (200,85) rectangle ++(50,10) node[pos=.5,black] {Hidden Units};
\draw [white] (200,65) rectangle ++(50,10) node[pos=.5,black] {Convolution};
\draw [white] (200,40) rectangle ++(50,10) node[pos=.5,black] {Input Array};

\end{tikzpicture}
\end{center}
\caption{The architecture applied to the rule learning task 
of \cite{marcusetal1999}, consisting of convolution followed by max-pooling 
and softmax. 
In the figure, the $12 \times 3$ grid at the bottom represents the 12 words within the 3 time-steps, with the first word at the top and the last at the bottom. 
This input is convolved with the two filters, 
resulting in the $12 \times 2$ output above it, 
and max-pooling reduces this to a single pair of values,
which are the logits of the output softmax.
In the example, \emph{wo fe wo} is encoded at the input as ones 
in the first and third channels for the syllable \emph{wo} 
and in the second channel for the syllable \emph{fe}. 
The same two filters are applied to the three channels of every word, 
with the $1 0 1$ filter matching the pattern of activations 
in the \emph{wo} position, 
giving a high activation value in the bottom channel 
at the same position in the hidden units. 
Max-pooling picks this value out and uses it to predict 
that this is an ABA sequence.
}\label{rl_fig}
\end{figure*}
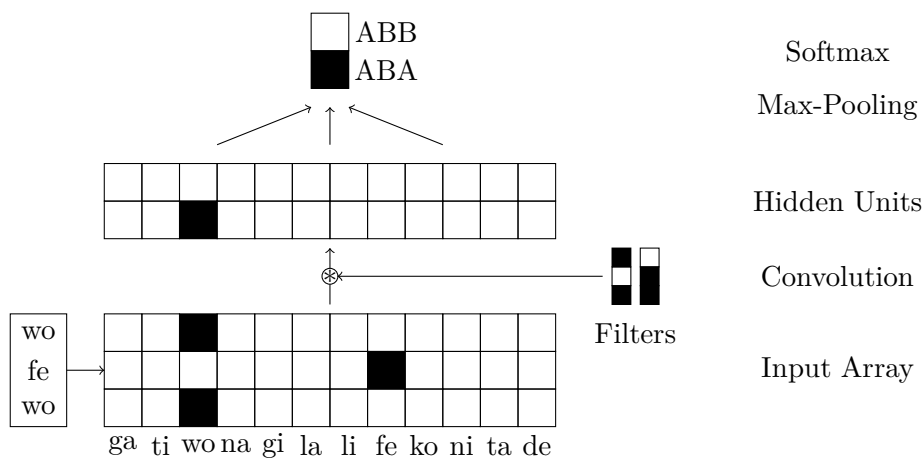

In this experiment we again compare two architectures, 
with and without the symmetry constraint. 
The former is a two layer network with 24 hidden units, 
consisting of a convolutional layer, 
followed by max-pooling
and softmax activations in the output layer. 
The latter is the same network 
with a fully connected layer replacing the 
initial convolutional layer. 
Again, the set of functions computable by the convolutional 
architecture is a proper subset 
of those computable by the unconstrained network.

Each data point consist of a three word sequence 
having the form ABB or ABA. 
These inputs can be thought of in terms of a $12 \times 3$ 
array, representing the twelve words in the vocabulary 
and the three time steps in the sequence, 
and the outputs are a binary pair of values, 
representing the two structures. 
In other words, for each time step, 
there is a one hot vector representing 
the word that occurs there. 
The training data contains 32 sequences, 
evenly balanced across structures, 
while the test data contains 4 sequences, 
consisting of 2 of each form. 
As discussed, the vocabulary of the test sequences 
is entirely disjoint from the training vocabulary, 
and as a result, training and test data sets lack 
any common active input units.

The convolutional architecture, shown in Figure \ref{rl_fig}, 
treats the three time steps 
as channels associated with each word,
to which it applies a filter of width one, 
sharing the same weights between all words. 
These filters project the three input channels 
down to two hidden layer channels for each word, 
corresponding to the two sequence structures. 
Max-pooling then passes the maximum value in each channel 
across the width of the hidden layer to the output units. 

The unconstrained network takes the same 
12 word $\times$ 3 time step matrix as its input, 
which it projects down onto 24 hidden units using a 
standard fully connected layer. 
As before, max-pooling is applied to these values to 
produce the two output values. 
In this way, only the word substitution symmetry constraint 
on the first layer changes between the two architectures.

The experiments are implemented in Tensorflow \citep{abadi}, 
using batch gradient descent over the full training set 
of 32 sequences.
The cross-entropy loss function is minimised 
for 1,000 epochs, and then performance is evaluated 
over the entire training and test sets. 
Accuracy is measured by discretising the outputs 
using a cutoff of 0.5, and an average is taken over 
100 random initialisations. 
A training run is restarted if the network fails to achieve 
100\% accuracy on the training set within 1,000 epochs.

\begin{table}[t]
  \begin{center}
  \begin{tabular}{lrr}
    \toprule
    
    Architecture & Training Accuracy &  Test Accuracy \\
    \midrule 
    Unconstrained & 100\% &  45\% \\
    Convolutional & 100\% & 100\% \\
    \bottomrule
  \end{tabular}
  \caption{Accuracies on the training and test sets of the rule learning task, with and without the constraint of word substitution symmetry.}\label{rl_res}
  \end{center}
\end{table}

The results in Table \ref{rl_res} demonstrate that, 
whereas the unconstrained network 
performs at the level of random chance on
the novel words in the test set, 
the convolutional architecture handles these sequences 
accurately. 
The first column shows that both models successfully learn 
to identify the structure of the sequences in the training set. 
However, performance of the two models diverges 
on the test set, as can be seen in the second column. 
The standard, unconstrained, multi-layer perceptron is unable 
to generalise its knowledge to the words in the test set 
because they involve independent sets of weights. 
In contrast, the convolutional architecture generalises 
effectively because the weights associated with the unseen 
words are shared with the seen words.

\subsection{Discussion}

Our convolutional network handled the test sequences 
effectively, despite their being composed of unseen word types. 
This, again, illustrates the ability of these architectures 
to generalise to inputs that lack featural overlap 
with the training data. 
In this case, the task required a more subtle analysis 
than for the copying of digits.
Specifically, we reasoned that if the outputs were to 
represent the structure of the inputs, disentangled 
from the particular words used, then they should be 
invariant to word substitutions that preserve 
the structure. 

Imposing that symmetry constraint on the network 
enabled it to recognise the \emph{same} structure 
when it was composed of new elements. 
Thus, our experiments have shown the relevance of symmetry 
in obtaining generalisation without featural overlap 
in two distinct situations: 
handling the same digit wherever it occurs and 
handling the same structure whichever words it contains. 
In addition to demonstrating the capacity of 
artificial neural networks to generalise 
in the absence of featural similarities, 
these results also raise the question 
of whether an analysis in terms of symmetry under 
transformations is also relevant to human cognitive abilities.

\section{General Discussion}

The role of similarity in the process of applying 
the lessons of past experience to new situations 
has been a longstanding topic of discussion. 
For example, \cite{hume} asserts that \emph{In reality, all arguments from experience are founded on the similarity which we discover among natural objects, and by which we are induced to expect effects similar to those which we have found to follow from such objects}. 
In psychology, this connection between similarity 
and inductive inferences has been explored 
in relation to verbal reasoning by 
\cite{rips}, \cite{osherson}, \cite{heit} and others. 
Taking a broader view, \cite{shepard} found that across a 
range of animals and with a variety of stimuli, 
generalisation could be modelled in terms of similarity 
within an abstract space.

In practical terms, this concept of generalisation as being 
grounded in relations of similarity underlies a variety of 
machine learning techniques such as 
k-nearest-neighbour classification \citep{fixhodges1951} 
and kernel machines \citep{aizerman}. 
As previously discussed, the view that 
the abilities of artificial neural networks 
are also limited to generalisation between similar items, 
in the sense of sharing overlapping features, 
has been a common assumption among both 
supporters and critics of connectionist modelling 
\citep{mcclellandplaut1999,marcus2001,alhamazuidema2019}. 

However, our experiments showed that convolutional architectures 
are not limited in this way, and are capable of generalising 
to test data that shares no features with the training data. 
Moreover, the underlying principles of this approach arose 
from models of visual cortex \citep{neocognitron}, 
and have become standard components in image processing 
networks \citep{alexnet,vgg,effnet}.
In concrete terms, weight sharing allows what is learned about 
features present in the training data to be transferred to 
unseen features in the test data. 

More abstractly, what connects the unseen test items to those 
seen in the training data is not simple featural similarity 
but instead a transformation that is a symmetry of the system. 
In the case of vision, the relevant transformation 
is spatial translation, allowing generalisation across 
positions within images. 
This transferred fairly directly to the identity learning task, 
where the test data contained familiar digits in new positions. 
In the case of the rule learning task, however, 
the relevant transformation was substitution 
of one word with another. 
Making the network symmetric under these substitutions 
allowed it to generalise from seen to unseen words. 

In fact, this idea that input stimuli can be connected not only 
by sharing common features, but also in terms of transformations 
that map representations onto each other arises in at least 
a couple of other domains. 
In analogical reasoning, for example, knowledge from a familiar 
base domain is leveraged in a novel target domain by identifying 
a mapping from one to the other \citep[e.g.][]{gentner}. 
Thus, analogy overcomes the limits of simple featural similarity 
in a comparable manner to that being discussed here in regards 
to artificial neural networks. 
In addition, transformations are frequently invoked to 
connect distinct grammatical structures in natural languages. 
So, passivisation - e.g. from \emph{John loves Mary} to 
\emph{Mary is loved by John} - is a transformation that 
preserves the basic semantics of sentences. 
\cite{domingos} have proposed an approach to semantic parsing 
based on treating such transformations as symmetries. 

Thus, the ideas that \emph{sameness} extends beyond simple 
featural overlap, and that it can be useful to identify 
transformations which connect distinct inputs 
are not novel within the cognitive sciences. 
Here, we have shown that this perspective can be effective  
in obtaining generalisation to novel features 
with artificial neural networks. 
In particular, we employed two types of symmetry, 
one relating to the translation of digits 
from one position to another, 
and another relating to the substitution of one 
word with another. 
More generally, we speculate that similar symmetries may be relevant 
in handling any structures in which multiple elements can be 
combined in multiple configurations, 
with the former allowing the same element to be treated 
consistently wherever it occurs, and the latter 
supporting recognition of the same structure  
whichever elements instantiate it.

In the identity learning task, 
translational symmetry ensured that the digits in all 
positions were handled consistently, 
resulting in the same copying operation being 
applied to the unseen digit position. 
Without this constraint, the behaviour of an item in a given 
position would be independent of its behaviour in any other, 
and therefore lack a notion of being the same item. 
A similar problem arises whenever an element can 
play multiple roles within a larger structure, 
for example \emph{John} in \emph{John loves Mary} 
and \emph{Alice loves John}.

In the rule learning task, 
symmetry under word substitutions allowed the 
ABB and ABA structures to be recognised 
irrespective of which words they contained.
In fact, such a symmetry will typically be required 
whenever a process needs to be driven by structure, 
in a way that is consistent whatever content instantiates 
that structure. 
For example, a valid logical inference - 
e.g. \emph{All men are mortal and Socrates is a man, 
therefore Socrates is mortal} - 
remains valid under substitutions - 
e.g. \emph{All men are mortal and Aristotle is a man, 
therefore Aristotle is mortal} - 
that preserve the form of the inference. 
Thus, the fact that validity is driven by formal structure 
implies that these substitutions are symmetries of the 
processes of logical derivation. 

The fact that the ABA and ABB structures containing both 
seen and unseen words are handled effectively implies that 
the knowledge embodied in the connection weights has been, 
to some extent, abstracted away from the particular 
examples seen during training.
\cite{marcusetal1999} interpret the behaviour of the infants 
in their experiments
as similarly implicating some form of abstract knowledge.
In particular, they suggest that the infants in their 
experiment had learned \emph{algebra-like rules
that represent relationships between placeholders (variables),
such as `the first item X is the same as the third item Y'}. 
While this mechanism for representing the structure of the 
sequences, in terms of using variables, is clearly distinct 
from the convolutional approach employed here, 
there are, nonetheless, underlying similarities. 
When, for instance, the network processes the sequence 
\emph{wo fe wo}, the \emph{wo} channels contain the values 
$101$, representing the fact that this particular word 
occurs at the beginning and end of the sequence. 
Moreover, from the point of view of filters 
that do not discriminate between words, 
these values simply signal that the same item 
occurs in both first and third position, 
which is what \cite{marcusetal1999} suggest the infants 
are sensitive to. 

In fact, we can think of the algebra-like rules and 
the weight sharing as two ways to implement 
the same requirement of symmetry under word substitution. 
In the algebraic case, the behaviour of the rules 
is invariant when the words seen during training 
are replaced with the unseen test words because 
the use of variables allows the rule to be expressed 
in a single form that makes no reference to any specific word, 
and is instead valid for all values. 
In contrast, the convolutional architecture replicates 
a version of the rule for every word, and in this way 
achieves invariance across values. 

Whatever the advantages and benefits of algebraic 
models of cognition, our concern here is with the 
abilities of artificial neural networks. 
In particular, our results showed that a convolutional 
architecture is capable of generalising to the 
unseen words and digits in the rule and identity learning tasks 
without the need for shared features in training 
and test inputs.

\section{Conclusions}

\cite{marcus2001} correctly identifies 
\emph{training independence} as a factor limiting 
the ability of standard connectionist architectures 
to generalise outside their training space. 
That is, without additional constraints, the backpropagation
algorithm adjusts the weights associated 
with each input feature independently. 
This flexibility makes them one of the most effective 
methods of machine learning when training data is plentiful 
and the test data is drawn from an identical distribution.
However, in the identity \citep{marcus1998} and 
rule \citep{marcusetal1999} learning tasks, 
this results in these models failing to generalise 
to the unseen features. 

Nonetheless, our experiments demonstrated that convolutional 
architectures can overcome this shortcoming, 
simply by sharing weights between seen and unseen features. 
This mechanism directly addresses the 
\emph{training independence} critique 
by tying the weights associated with multiple features.
More abstractly, the particular form of weight sharing 
was driven by considerations 
of relevant symmetries identified in each task. 
This approach allowed us to extend a notion of sameness to 
inputs that shared no common features, 
and as a consequence the trained networks were able to 
apply the same rule to an unseen digit position and 
recognise the same structure instantiated with novel elements. 

\section*{Acknowledgements}

This research was supported by the European Research Council Grant Generalization in Mind and Machine, ID number 741134.

\appendix
\section{Convolutional Neural Networks}\label{app_cnn}

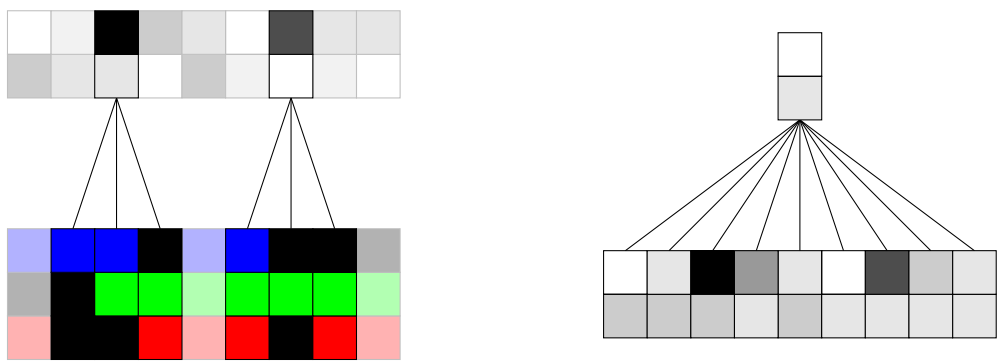
\begin{figure*}[t]
  \centering
  \begin{subfigure}[t]{0.48\textwidth}
  \centering
\begin{tikzpicture}[draw=black!50, node distance=\layersep, scale=0.29]

\draw [lightgray,fill=red!30] (0,0) rectangle (2,2);
\draw [lightgray,fill=black!30] (0,2) rectangle (2,4);
\draw [lightgray,fill=blue!30] (0,4) rectangle (2,6);

\draw [lightgray,fill=red!30] (8,0) rectangle (10,2);
\draw [lightgray,fill=green!30] (8,2) rectangle (10,4);
\draw [lightgray,fill=blue!30] (8,4) rectangle (10,6);

\draw [lightgray,fill=red!30] (16,0) rectangle (18,2);
\draw [lightgray,fill=green!30] (16,2) rectangle (18,4);
\draw [lightgray,fill=black!30] (16,4) rectangle (18,6);

\draw [black,fill=black] (2,0) rectangle (4,2);
\draw [black,fill=black] (2,2) rectangle (4,4);
\draw [black,fill=blue] (2,4) rectangle (4,6);
\draw [black,fill=black] (4,0) rectangle (6,2);
\draw [black,fill=green] (4,2) rectangle (6,4);
\draw [black,fill=blue] (4,4) rectangle (6,6);
\draw [black,fill=red] (6,0) rectangle (8,2);
\draw [black,fill=green] (6,2) rectangle (8,4);
\draw [black,fill=black] (6,4) rectangle (8,6);

\draw [black,fill=red] (10,0) rectangle (12,2);
\draw [black,fill=green] (10,2) rectangle (12,4);
\draw [black,fill=blue] (10,4) rectangle (12,6);
\draw [black,fill=black] (12,0) rectangle (14,2);
\draw [black,fill=green] (12,2) rectangle (14,4);
\draw [black,fill=black] (12,4) rectangle (14,6);
\draw [black,fill=red] (14,0) rectangle (16,2);
\draw [black,fill=green] (14,2) rectangle (16,4);
\draw [black,fill=black] (14,4) rectangle (16,6);

\draw [black] plot coordinates {(5,6) (5,12)};
\draw [black] plot coordinates {(3,6) (5,12)};
\draw [black] plot coordinates {(7,6) (5,12)};

\draw [black] plot coordinates {(13,6) (13,12)};
\draw [black] plot coordinates {(11,6) (13,12)};
\draw [black] plot coordinates {(15,6) (13,12)};

\draw [lightgray,fill=black!20] (0,12) rectangle (2,14);
\draw [lightgray] (0,14) rectangle (2,16);
\draw [lightgray,fill=black!10] (2,12) rectangle (4,14);
\draw [lightgray,fill=black!5] (2,14) rectangle (4,16);

\draw [lightgray] (6,12) rectangle (8,14);
\draw [lightgray,fill=black!20] (6,14) rectangle (8,16);
\draw [lightgray,fill=black!20] (8,12) rectangle (10,14);
\draw [lightgray,fill=black!10] (8,14) rectangle (10,16);
\draw [lightgray,fill=black!5] (10,12) rectangle (12,14);
\draw [lightgray] (10,14) rectangle (12,16);

\draw [lightgray,fill=black!5] (14,12) rectangle (16,14);
\draw [lightgray,fill=black!10] (14,14) rectangle (16,16);
\draw [lightgray] (16,12) rectangle (18,14);
\draw [lightgray,fill=black!10] (16,14) rectangle (18,16);

\draw [black,fill=black!10] (4,12) rectangle (6,14);
\draw [black,fill=black] (4,14) rectangle (6,16);

\draw [black] (12,12) rectangle (14,14);
\draw [black,fill=black!70] (12,14) rectangle (14,16);

\end{tikzpicture}
\caption{A convolutional layer. The same learned function is applied to each small patch of an image, by sharing the same weights at all positions. Two such patches and their shared weights are highlighted.}\label{conv_fig}
\end{subfigure}
\hfill
  \begin{subfigure}[t]{0.48\textwidth}
  \centering
\begin{tikzpicture}[draw=black!50, node distance=\layersep, scale=0.29]

\draw [white] (0,1) rectangle (2,2);

\draw [black,fill=black!20] (0,2) rectangle (2,4);
\draw [black] (0,4) rectangle (2,6);

\draw [black,fill=black!20] (8,2) rectangle (10,4);
\draw [black,fill=black!10] (8,4) rectangle (10,6);

\draw [black,fill=black!10] (16,2) rectangle (18,4);
\draw [black,fill=black!10] (16,4) rectangle (18,6);

\draw [black,fill=black!20] (2,2) rectangle (4,4);
\draw [black,fill=black!10] (2,4) rectangle (4,6);
\draw [black,fill=black!20] (4,2) rectangle (6,4);
\draw [black,fill=black] (4,4) rectangle (6,6);
\draw [black,fill=black!10] (6,2) rectangle (8,4);
\draw [black,fill=black!40] (6,4) rectangle (8,6);

\draw [black,fill=black!10] (10,2) rectangle (12,4);
\draw [black] (10,4) rectangle (12,6);
\draw [black,fill=black!10] (12,2) rectangle (14,4);
\draw [black,fill=black!70] (12,4) rectangle (14,6);
\draw [black,fill=black!10] (14,2) rectangle (16,4);
\draw [black,fill=black!20] (14,4) rectangle (16,6);

\draw [black] plot coordinates {(1,6) (9,12)};
\draw [black] plot coordinates {(3,6) (9,12)};
\draw [black] plot coordinates {(5,6) (9,12)};
\draw [black] plot coordinates {(7,6) (9,12)};
\draw [black] plot coordinates {(9,6) (9,12)};
\draw [black] plot coordinates {(11,6) (9,12)};
\draw [black] plot coordinates {(13,6) (9,12)};
\draw [black] plot coordinates {(15,6) (9,12)};
\draw [black] plot coordinates {(17,6) (9,12)};

\draw [black,fill=black!10] (8,12) rectangle (10,14);
\draw [black] (8,14) rectangle (10,16);

\end{tikzpicture}
\caption{A global pooling layer. A fixed translationally invariant function is applied to all the units in each channel of the input layer to produce a single value in each channel of the output. }\label{pool_fig}
\end{subfigure}
\caption{The structure of convolutional and global pooling layers}
\end{figure*}

Many of the mechanisms found in modern 
Convolutional Neural Networks have their origins in 
the Neocognitron \citep{neocognitron}. 
This model consisted of multiple layers of 
units inspired by the simple and complex cells 
described by \cite{hubelwiesel}. 
Local features were learned by S-cells 
which, crucially, share their parameters 
across the input array. 
This creates a field of replicated feature detectors, 
which are then aggregated locally by C-cells, 
creating wider detectors with limited shift invariance. 
In this way, the receptive fields of these units grow 
as processing proceeds through the layers of the network, 
building more complex features and expanding the 
tolerance to spatial translation.

The sharing of feature weights across the visual field 
within this model is critical in its achieving 
invariance under spatial translation. 
Without it, different features would be learned 
in different positions, preventing a consistent 
response to the same object wherever it occurs. 
It is unclear whether some form of this mechanism can be made 
biologically plausible, but there is nonetheless 
evidence that human object recognition is tolerant 
to substantial displacements across the retina \citep{blything}. 
Moreover, \cite{marcus2018} includes translation invariance 
in a list of 10 computational primitives that he thinks 
would need to be part of the innate machinery 
supporting any human-like cognition.

Beyond the Neocognitron, weight sharing has been a common 
tool for achieving the desired form of generalisation 
in other visual applications \citep[e.g.,][]{rumelhartpdp}. 
\cite{rumelhartnature} also explained generalisation 
within a relational learning task in terms of shared weights. 
Similarly, \cite{jain} imposed weight sharing on their network 
in order to obtain appropriate generalisation in a 
model of natural language parsing.

The same ideas also found application in practical tasks, 
such as handwritten digit recognition \citep{lenet}, 
where robust generalisation to small deformations was vital. 
Despite these strengths and the reduced number of free parameters,  
effective training of such models still required 
large amounts of data, which motivated the development of 
datasets such as ImageNet \citep{imagenet}. 
Given these extensive collections of natural images, 
it has become possible to train neural models of object recognition 
with impressive test set performance 
\citep[e.g.,][]{alexnet,vgg,effnet}. 

These models are based, mainly, around the ideas introduced 
in the Neocognitron \citep{neocognitron}, 
consisting of layers of identical feature detectors 
combined with layers of units that pool these features 
into larger structures. 
The general term \emph{convolutional neural network} is used to 
refer to the various variations on this design, 
in reference to the mathematical name for an operation equivalent 
to applying the same filter to all positions in an image. 
The convolutional layers themselves contain the trainable 
feature detectors, and these are typically followed by 
pooling layers which apply a fixed aggregation to those outputs, 
such as taking an average or maximum.
Particular architectures in this family may combine these 
convolution and pooling layers with other components 
such as fully connected layers or skip-connections.

A convolutional layer is depicted in more detail in 
Figure \ref{conv_fig}. 
An input image is represented by the $8 \times 3$ grid 
at the bottom, with spatial position depicted horizontally 
and the red, green and blue colour channels vertically. 
For simplicity, only one spatial dimension is shown, 
but the extension to two dimensions is straightforward. 
Each unit in the output layer is driven by a small patch 
in the input, and two such patches are highlighted.
In this case, there are two output channels, so, given the 
three input channels and spatial width of three, 
the number of weights used in this transformation is 
$2 \times 3 \times 3 = 18$. 
In a real image processing model, there would, of course, 
be an additional spatial dimension and typically 
a larger number of output channels is used, 
but the fundamental principle remains that 
the same weights are applied to all positions in the input. 
This constraint ensures that, under spatial translations $T$, 
the function, $conv(x)$, computed by the layer 
is \emph{equivariant}: $conv(T x) = T conv(x)$.

Figure \ref{pool_fig} depicts a global pooling layer, 
applied to a two channel input. 
In this case, a fixed translationally invariant function, 
such as taking the maximum or average, is applied to 
the features across all the input positions
within each channel. 
This preserves the number of input channels, 
but reduces the information across the input positions 
down into a single spatially \emph{invariant} stream. 
That is, the function is insensitive to input translations: 
$pool(x) = pool(T x)$.
Typically, a convolutional architecture will also contain 
local pooling layers, which aggregate across receptive fields 
of limited extent, comparable to the complex cells 
of the mammalian visual system. 
These are then interleaved with convolutional layers 
to produce larger and more complex feature detectors 
as processing proceeds through the network. 
Finally, a global pooling layer can be used to 
aggregate across the entire image in the final layers 
of the network.
Alternatively, a fully-connected layer is sometimes 
used instead, but this tends to reduce the translational 
invariance of the network \citep{blything}.

In practice, a number of other factors, 
related to training data and network design, 
may also obstruct the theoretical ideal 
of perfect translation invariance \citep{azulayweiss2019}. 
While humans can typically recognise an object anywhere 
in their visual field after seeing it in a single position,
the networks applied to image recognition tasks 
frequently lack such flexible generalisation abilities
\citep{biscionebowers2020}.
Nonetheless, a number of computational neuroscientists 
see appropriately trained convolutional architectures 
as having a close resemblance to the primate visual system 
\citep[e.g.][]{yaminsetal2014,kriegeskorte2015,kubiliusetal2016}. 

\bibliography{main}
\bibliographystyle{apalike}

\end{document}